# MGFs: Masked Gaussian Fields for Meshing Building based on Multi-View Images

Tengfei Wang, Zongqian Zhan*, Rui Xia, Linxia Ji, Xin Wang*

*School of Geodesy and Geomatics, Wuhan University, 129 Luoyu Road, Wuhan 430072, People's Republic of China*

A B S T R A C T

Over the last few decades, image-based building surface reconstruction has garnered substantial research interest and has been applied across various fields, such as heritage preservation, architectural planning, etc. Compared to the traditional photogrammetric and NeRF-based solutions, recently, Gaussian fields-based methods have exhibited significant potential in generating surface meshes due to their time-efficient training and detailed 3D information preservation. However, most gaussian fields-based methods are trained with all image pixels, encompassing building and nonbuilding areas, which results in a significant noise for building meshes and degeneration in time efficiency. This paper proposes a novel framework, Masked Gaussian Fields (MGFs), designed to generate accurate surface reconstruction for building in a time-efficient way. The framework first applies EfficientSAM and COLMAP to generate multi-level masks of building and the corresponding masked point clouds. Subsequently, the masked gaussian fields are trained by integrating two innovative losses: a multi-level perceptual masked loss focused on constructing building regions and a boundary loss aimed at enhancing the details of the boundaries between different masks. Finally, we improve the tetrahedral surface mesh extraction method based on the masked gaussian spheres. Comprehensive experiments on UAV images demonstrate that, compared to the traditional method and several NeRF-based and Gaussian-based SOTA solutions, our approach significantly improves both the accuracy and efficiency of building surface reconstruction. Notably, as a byproduct, there is an additional gain in the novel view synthesis of building.

## 1. Introduction

Accurate and realistic digital reconstruction of building, including surface meshes and novel view synthesis, serves as a critical foundation for applications in smart cities, virtual reality, and disaster analysis. Currently, two common data sources, airborne LiDAR and images from Unmanned Aerial Vehicles (UAVs), are widely used in building reconstruction (Zhang et al., 2024). However, the acquisition of airborne LiDAR data is costly and often suffers from incomplete facade representations. Moreover, the lack of texture makes novel view synthesis inherently unfeasible. In contrast, imagery has emerged as a prevalent and suitable data source for building digital reconstruction due to its low acquisition cost and ease of operation.

Traditional photogrammetric methods for dealing with images have obtained ample achievements, as evidenced by many well-established packages: Metashape (Agisoft, 2022), ContextCapture (Bentley, 2018), COLMAP (Schonberger and Frahm, 2016), MicMac (Pierrot Deseilligny and Clery, 2012). Their processing workflow typically involve,s four steps: *First*, Structure-from-Motion (SfM) (Özyeşil et al., 2017) or image orientation (Wang et al., 2019; X. Wang et al., 2021), which aims to solve image poses and sparse 3D point cloud; *Second*, Multi-View Stereo (MVS) (Stathopoulou and Remondino, 2023), encompassing dense matching and point cloud fusion; *Third*, mesh extraction based on the point cloud, involving initial mesh reconstruction and mesh optimization (Newman and Yi, 2006; Kazhdan and Hoppe, 2013); (4) Texture mapping (Dostal and Yamafune, 2018), assigning each triangle a texture with realistic color. The whole pipeline is cumbersome and time-consuming, often resulting in reconstructions with holes, missing details, and redundant polygons. To address these limitations, several learning-based dense reconstruction methods, such as MVSNet (Yao et al., 2018) and TransMVSNet (Ding et al., 2022), have been proposed, achieving competitive performance. These methods integrate various steps of the traditional pipeline into an end-to-end learnable network, thereby simplifying the workflow. However, they require large amounts of training data and incur high computational costs.

Moreover, the aforementioned methods primarily focus on synthesizing surface meshes and are not well-suited for accurate rendering and virtual display of entire building. In recent years, 3D reconstruction methods based on Neural Radiance Fields (NeRF) (Mildenhall et al., 2020), such as NeuS (P. Wang et al., 2021) and VoLSDF (Yariv et al., 2021), have shown promise in addressing these issues. These methods can generate detailed surface models and perform novel view rendering. However, they often suffer from long training times, low rendering efficiency, and limited capability to handle outdoor scenes with numerous high-resolution images.

---

*   Corresponding author.

    E-mail address: zqzhan@sgg.whu.edu.cn; xwang@sgg.whu.edu.cn.

Recently, to accelerate the training and rendering, 3D Gaussian Splatting (3DGS) (Kerbl et al., 2023) has initiated a new research hotspot in the field of novel view rendering. The 3DGS combines the advantages of implicit radiance fields and traditional explicit point cloud representations, using multi-dimensional Gaussian spheres to reconstruct scenes. In terms of rendering, the Gaussian spheres are splatted onto the 2D image plane to integrate the color information, resulting in a rendering speed that is tens of times faster than normal NeRF. Inspired by 3DGS, some relevant surface reconstruction methods, such as SuGaR (Guédon and Lepetit, 2024) and GOF (Yu et al., 2024b), have also been developed and achieved remarkable results. However, when meshing building based on UAV images, these methods are often adversely affected by the presence of non-building regions. More specifically, the gaussian spheres in non-building regions may introduce noise when rendering and meshing building. Moreover, the training time is undesirably increased by the computation of the rays passing through non-building pixels.

To enhance the performance of Gaussian fields-based methods for meshing building from multi-view images, we propose a new framework called Masked Gaussian Fields (MGFs). First, for a small subset of images, a state-of-the-art (SOTA) segmentation method, EfficientSAM (Xiong et al., 2024), is applied to efficiently generate masks for entire building. Additionally, multi-level masks with different level of details are predicted as well. Subsequently, the estimated internal and external orientation parameters, derived from COLMAP, are used to automatically generate corresponding masks for the remaining images and to label sparse points belonging to building. Second, using these masked images and points, the masked Gaussian fields are trained by incorporating two innovative losses: a multi-level perceptual masked loss, which focuses on building regions, and a boundary loss, which concentrates on enhancing the detail of boundaries between different masks. Finally, an improved tetrahedron extraction method based on masked gaussian spheres and multi-directional screening is presented to extract building mesh. As a byproduct, the rendered novel views of building can be generated with improved quality after applying the proposed method (see section 4.2). In general, our contributions are *threefold*:

1) **Generation of Masked Gaussian Fields**. To differentiate the regions of building and non-building, we utilize EfficientSAM along with COLMAP to efficiently generate masks for Gaussian fields, wherein an entire mask for the whole building region and multi-level masks for sub-elements of the building are generated. Only the image pixels and sparse points masked as building are used for subsequent training.
2) **Boundary loss and multi-level perceptual masked loss**. To enhance details of boundaries between various masks, a novel boundary loss is estimated using the predicted values of boundary rays based on a new weighted volume rendering formula. Additionally, a multi-level perceptual masked loss, encompassing all the pixels belonging to the detected multi-level masks, is employed to ensure local consistency within each mask.
3) **Improved mesh extraction method**. To tackle the limitation of slow and redundant mesh extraction, we propose a tetrahedral mesh extraction strategy with multi-directional screening based on masked information.

## 2. Related work

In this section, we first review some relevant works on surface reconstruction using multi-view images, and then followed by a closely related task of novel view rendering containing volume density optimization, which can be further employed for generating surface meshes.

### 2.1. Surface reconstruction of building based on multi-view images

Over the past few decades, considerable efforts have been made in surface reconstruction using multi-view images. There are three main categories of methods: traditional solutions, learning-based approaches, and volume-rendering-based methods.

**Traditional.** Referring to Seitz et al. (2006), traditional multi-view reconstruction methods can typically be divided into four categories: (1) volume-based; (2) surface evolution-based; (3) feature point-based; (4) depth map-based. Voxel-based methods typically compute a cost function over a 3D volume and extract surfaces from the voxels. However, these methods are always limited by high memory cost and voxel resolution, making them infeasible for complex and large scenes (Romanoni et al., 2017; Savinov et al., 2016). Surface evolution-based methods iteratively evolve surfaces via adding or removing elements to minimize an energy function, whereas their time efficiency is highly dependent on the initialization (Heise et al., 2015; Li et al., 2016). Feature point-based methods first extract and match a set of feature points, directly fit surfaces using these points. However, relying solely on feature points cannot guarantee the reconstruction accuracy of the whole scene (Locher et al., 2016; Wu et al., 2010). Depth map-based methods are widely used due to their high flexibility as they convert the 3D reconstruction problem into depth map estimation in 2D space (Shen, 2013; Xu and Tao, 2019).

**Learning-based**. Learning-based MVS methods use neural networks to estimate image similarity, replacing manually designed features, and have developed end-to-end pipelines that simplify the steps of mesh reconstruction (Wu et al., 2024). However, these learning-based multi-view reconstruction method typically require substantial amounts of real-world data for supervised training, which limits their generalization. In addition, researchers have proposed unsupervised methods that primarily rely on photometric consistency (Chang et al., 2022; Khot et al., 2019). Although these methods do not require training labels, the reconstruction accuracy is generally inferior to supervised methods.

**Volume rendering-based**. In recent years, volume rendering has occupied an essential position in geometric surface reconstruction. For example, NeRF-based methods have become prominent due to their superior performance on small scenes or desktop-sized toys (Mildenhall et al., 2021). These methods can generate meshes end-to-end using only camera poses and raw images as input (Li et al., 2023; Chen et al., 2024). While these methods can extract relatively accurate surface meshes, they are resource-intensive, requiring high GPU memory and extensive training time, which makes them difficult to apply to large outdoor scenes.

In the last year, the emergence of another volume rendering method, 3D Gaussian Spheres (3DGS) (Kerbl et al., 2023), has provided new insights. Its explicit 3D representation has shown significant advantages in editability and training speed compared to NeRF-based methods that require implicit representation. So far, based on 3DGS, many surface reconstruction methods have been proposed. NeuSG (Chen et al., 2023) and DN-splatter (Turkulainen et al., 2024) both use additional supervision information as constraints; the former uses normals generated by NeuS as additional supervision, and the latter uses a monocular depth estimation network (Bhat et al., 2023) to estimate depth. Although these methods can achieve high-precision meshes, some extra information are prerequired to be generated for supervised training. In contrast, Guédon et al. (2024) proposed SuGaR, which adopts an unsupervised regularization, to force the alignment of Gaussian spheres with the scene, but its surface reconstruction performanceis relatively poor. More recently, 3DGS has been further improved by introducing different volume rendering methods for surface reconstruction such as 2DGS and GOF, etc. 2DGS (Huang et al., 2024) generates 2D directional Gaussian discs which is applied for volume rendering and GOF (Yu et al., 2024b) proposes Gaussian opacity fields to estimate better level set. However, these methods do not consider the noise interference introduced by background information into the target area.

All the aforementioned multi-view surface reconstruction methods can be also applied to deal with building. For example, traditional methods have been used to reconstruct houses and bridges, as demonstrated by Hwang et al. (2016) and Yan et al. (2022), Ge et al. (2024) employed NeRF-based methods for the reconstruction and digital preservation of ancient building.

### 2.2. Novel view rendering

Novel view rendering, also known as novel view synthesis, typically involves generating multiple images of new viewpoints based on a set of existing images from known viewpoints. Conventional methods (Liu et al., 2019; Waechter et al., 2014) often commence with a surface mesh or voxel grid mapped with textures. These textured meshes are then back-projected onto the target image using known pose parameters. These approaches heavily rely on the textured mesh model, which generally requires extensive computation and significantly impacts the quality of the rendered novel views.

In recent years, Neural Radiance Fields (NeRF) have provided a novel solution for image rendering, capable of synthesizing high-fidelity novel views directly from 2D multi-view images and camera poses, without requiring explicit 3D representations like 3D mesh models. However, the multi-layer perceptron (MLP) architecture used in NeRF and the inherent integration complexity of predicting color information for each pixel result in prolonged training times and low frames per second (FPS) during rendering. To address this issue, Müller et al.(2022) proposed Instant-NGP, an effective solution utilizing multi-resolution hash encoding to improve neural network inputs, significantly reducing training time. Despite these improvements, Instant-NGP exhibits deficiencies in accuracy and frequency during rendering. Conversely, Mip-NeRF 360 (Barron et al., 2022) and Zip-NeRF (Barron et al., 2023) demonstrate superior rendering quality but still suffer from slow training and rendering speeds. Generally, NeRF-based rendering methods struggle to balance accuracy and efficiency.

In contrast to NeRF-based methods, 3D Gaussian Splatting (3DGS) has emerged as a new research hotspot. 3DGS represents a 3D scene using explicit Gaussian point clouds, enabling rapid rendering while maintaining high accuracy. The rendering speed is significantly faster than Instant-NGP, and the precision surpasses most existing NeRF methods (Chen and Wang, 2024). Additionally, recent efforts have focused on improving rendering performance: Lu et al. (2024) reduced redundant Gaussians by optimizing anchor point distribution, Yu et al. (2024a) enhanced rendering effects through anti-aliasing techniques, and various works have applied compression algorithms to achieve lightweight models, thereby saving memory during 3DGS rendering (Lee et al., 2024; Niedermayr et al., 2024).

## 3. Method

Fig. 1 illustrates our overall framework for surface reconstruction of building using images, comprising three successive parts: masked information generation, Masked Gaussian Fields training and rendering, and surface mesh extraction. In the first part, we introduce the automatic generation of masked information via combining COLMAP (Schonberger and Frahm, 2016) with EfficientSAM (Xiong et al., 2024), which includes the generation of multi-level masks and masked points of building (see more details in Section 3.1). Subsequently, in the second part, based on the masked information, the training and rendering process of Masked Gaussian Fields are presented (see more details in Section 3.2). Finally, based on the trained Gaussian masked fields, we perform multi-directional filtering on the Gaussian spheres to obtain the signed distance function (SDF). Then, the improved marching tetrahedron method is used to reconstruct the surface mesh (see more details in Section 3.3).

### 3.1. Masked information generation

This section outlines the methodology for acquiring multi-level masks and masked points using EfficientSAM and COLMAP, as depicted in the first part of Fig. 1.

#### 3.1.1 Camera pose calculation and sparse point cloud generation

Volume rendering methods require camera poses and corresponding sparse points as input. In this work, we use the well-established SfM package, COLMAP, to calculate the camera poses and generate a sparse point cloud for the entire scene. However, the sparse point cloud obtained by COLMAP is redundant, as it contains both building and non-building regions, which may affect the accuracy of subsequent reconstruction. In the following subsections, we will detail the process of generating masked points that include only the building regions.

#### 3.1.2 Automatic generation of multi-level masks and masked points

In this work, we apply the SOTA segmentation architecture, EfficientSAM, with the SAMI-Pretrained Image Encoder and Masked Decoder, to efficiently and accurately generate masked information. Although EfficientSAM is capable of fully automatic mask generation, we manually select a few prompted points to facilitate segmentation and further enhance its performance. To reduce the manual work, only a small subset of images is annotated with prompted points, and all the remaining images are totally automatically processed to generated masked information by leveraging the results of SfM. A detailed description of this process is provided below.

Based on the spatial distribution of input images which is derived from estimated poses, and then 1/5 of the images are selected as root images using uniform sampling, with the remaining images considered as child images. The root images are first input into the EfficientSAM, where prompted points are manually annotated for building segmentation, resulting in the corresponding building masks (called root masks). For the child images, the corresponding child masks are generated using camera poses and the collinearity Equation. The details are as follows:

Firstly, the root masks are generated by pre-trained EfficientSAM:

$$M_{root} = E_{SAM(I_{root})} \qquad (1)$$

$M_{root}$ represents the set of all root masks, $I_{root}$ represents the set of all root images, $E_{SAM}$ represents the mask segmentation operation by EfficientSAM.

Secondly, we back-project the 3D points obtained by COLMAP onto the masked root images, retaining only the 3D points whose 2D image

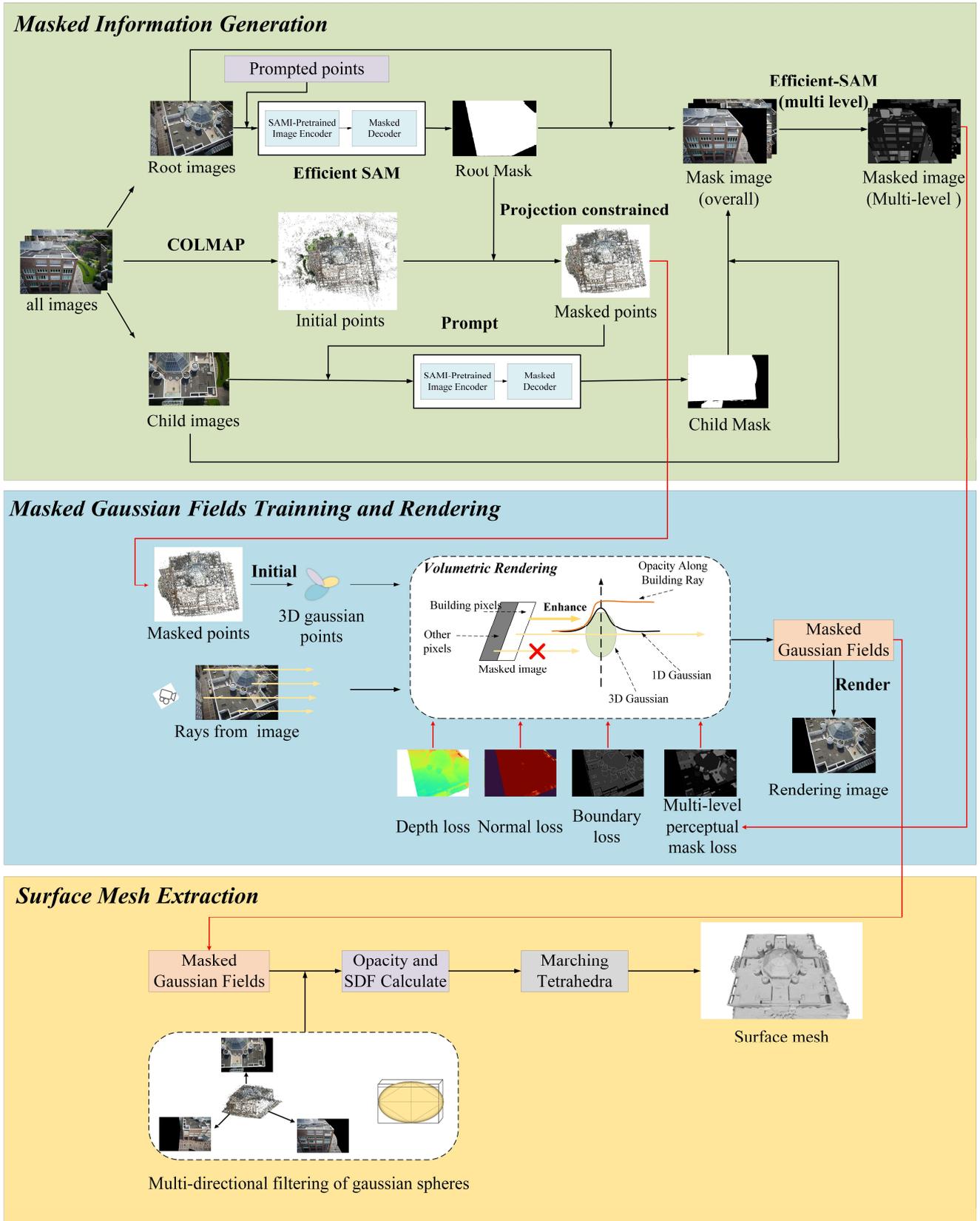

Fig. 1 Overall workflow of our MGFs for meshing surface of building using multi-view images

points coordinates fall within the root masks, as root masked points.

More specifically, we calculate the projection coordinates $p_{ij}$ of the 3D points $P_i$ on each root image according to Equation (2):

$$\begin{pmatrix} u_{ij} \\ v_{ij} \\ 1 \end{pmatrix} = K_j \left( R_j \begin{pmatrix} X_i \\ Y_i \\ Z_i \end{pmatrix} + t_j \right) \quad (2)$$

where, $P_i = (X_i, Y_i, Z_i)$ is the $i$-th 3D point, $K_j$ is the intrinsic matrix of the $j$-th image, $R_j$ and $t_j$ are the rotation matrix and translation vector of the $j$-th image, respectively. $p_{ij} = (u_{ij}, v_{ij})$ represents the back-projected coordinates of the $i$-th 3D point on the $j$-th image. If $p_{ij}$ is inside the root mask $M_j$, we can set $M_j(u_{ij}, v_{ij}) = 1$; otherwise, $M_j(u_{ij}, v_{ij}) = 0$.

For the set of 3D points inside the masks, $P'$ (i.e., the masked points), can be represented by the following formula:

$$P' = \{P_i \in P | \forall j, M_j(u_{ij}, v_{ij}) = 1\} \quad (3)$$

Then, according to Equation (2), we project the masked 3D points $P'$ onto the child images as new prompted points. These points are then used by EfficientSAM to generate child masks. Now, the building area segmentation for all images is completed, and the masked points are obtained.

For a given image, the above operation generates the first-level mask of the entire building. Then, EfficientSAM is employed again to generate a second-level mask containing more detailed elements, such as windows, doors, and walls, based on the corresponding pixel prompts for these elements. In this paper, the multi-level masks are composed of these various levels of masks, denoted as $M_{ms}$. Specifically, $M_{ms}(u, v) = 0$ if pixel $(u, v)$ is outside the detected masks; otherwise, $M_{ms}(u, v) \ne 0$.

### 3.2. Gaussian Masked Fields training and rendering

In this section, we elucidate the process of utilizing masked points and multi-level masked images obtained in Section 3.1 as inputs for training Masked Gaussian Fields and performing rendering operations. Initially, during the volume rendering process, we enhance the accuracy of color prediction by increasing the light density at the edges of the mask and filtering out light rays in non-building areas (Section 3.2.1). Subsequently, to refine the predicted colors and match them closely to the real colors, we introduce a multi-level mask-aware loss to improve local consistency. Additionally, we propose a boundary loss to strengthen the boundary information of building components, thereby achieving precise building rendering results (Section 3.2.2).

#### 3.2.1 Volume rendering of Gaussian Masked Fields

To make this paper more comprehensive, we provide a description of the definition of 3D Gaussian as outlined by Kerbl et al. (2023). The 3D Gaussian function $G(x)$ is defined as follows:

$$G(x) = \exp\left(-\frac{1}{2}(x-p)^T \Sigma^{-1}(x-p)\right) \quad (4)$$

where $p$ is the center of the Gaussian sphere, $\Sigma$ is the covariance matrix, usually indicated as a combination of the scaling matrix $S$ and the rotation matrix $R$:

$$\Sigma = RSS^T R^T \quad (5)$$

In this paper, we refer to the Gaussian Opacity Fields (GOF) (Yu et al., 2024b) to construct the Masked Gaussian Fields. For each Gaussian point $G_k(x)$, the original coordinate system is transformed into the Gaussian local coordinate system. The transformation of any point $x$ to $x_g$ in the local coordinate system is as follows:

$$o_g = (R_k(o - p_k)) \otimes s_k^{-1} \quad (6)$$

$$r_g = R_k r \otimes s_k^{-1} \quad (7)$$

$$x_g = o_g + t r_g \quad (8)$$

In this local coordinate system, the Gaussian value of any point along the ray is a one-dimensional Gaussian value $G_k^{1D}(t)$, calculated as follows:

$$G_k^{1D}(t) = G_k(x_g) = e^{-\frac{1}{2}x_g^T x_g} \quad (9)$$

When $t = t^*$, $G_k^{1D}(t)$ takes the maximum Gaussian value:

$$t^* = -\frac{r_g^T o_g}{r_g^T r_g} \quad (10)$$

Therefore, for a given camera center $o$ and light direction $r$, the contribution of the Gaussian sphere $G_k$ is defined as:

$$\mathcal{E}(G_k, o, r) = G_k^{1D}(t^*) \quad (11)$$

Considering the internal and edge structures of building on images, we combine the multi-level masks obtained in Section 3.1.2 with the Gaussian mask fields. In particular, the light density of the boundaries derived from the multi-layer masks is enhanced in the volume rendering, and ignores the light in the non-mask area.

In this paper, we propose a function to weight masks' boundaries. Firstly, the edge detection operator $K$ (convolution kernel) is used to obtain boundaries between multi-level masks $M_{ms}$, i.e., $M_{bd} = M_{ms} * K$, wherein $M_{bd} \ne 0$ if the corresponding pixel belongs to boundaries. The weight function $w(u, v)$ is defined as:

$$w(u, v) = \begin{cases} w_{edge} & \text{if } M_{ms}(u,v)\ne 0 \text{ and } M_{bd}\ne 0 \\ 1 & \text{if } M_{ms}(u,v)\ne 0 \text{ and } M_{bd}==0 \\ 0 & \text{if } M_{ms}(u,v)==0 \end{cases} \quad (12)$$

Here, $w_{edge}$ is a constant value (set as 10 in our work), indicating the influence of the boundaries in the masks. Base on this, the boundary-weighted contribution value of the Gaussian sphere $G_k$ is computed as::

$$\mathcal{E}^w(G_k, o, r) = w(u_{ij}, v_{ij}) \mathcal{E}(G_k, o, r) \quad (13)$$

To leave out the rays from non-mask region (i.e., non-building region for this paper), we define an indicator function $\chi$:

$$\chi(o, r) = \begin{cases} 0 & \text{if}(o,r) \text{ belongs to the mask area} \\ 1 & \text{otherwise} \end{cases} \quad (14)$$

Finally, an enhanced volume rendering formula is presented as:

$$C^w(o, r) = \chi(o, r) \sum_{k=1}^{K} c_k \alpha_k^w \mathcal{E}^w(G_k, o, r) \prod_{j=1}^{k-1}\left(1 - \alpha_j^w \mathcal{E}(G_j, o, r)\right) \quad (15)$$

where $\alpha_k, \alpha_j$ is the opacity of the $k$-th and $j$-th Gaussian sphere, $c_k$ is the corresponding color information of $G_k$.

#### 3.2.2 Training loss

To train a Gaussian fields, referring to 3DGS(Kerbl et al., 2023), a common loss function of RGB loss $Loss_{rgb}$ is widely used, defined as

$$Loss_{rgb} = \alpha \cdot Loss_{L1} + \beta \cdot Loss_{ssim} \quad (16)$$

$L_{loss}$ is the L1 loss. $Loss_{ssim}$ is the structural similarity index loss.

Based on Equation (16), this section explains our proposed two new training losses: boundary loss $Loss_{boundary}$, and the multi-level perceptual mask loss: $Loss_{mlpm}$.

**(1) Boundary loss**

For the building in images, boundaries are typically found at the intersections of various elements, such as windows and walls, and are characterized by significant changes in brightness, color, texture, and shading in the transition areas. Therefore, accurately identifying and reconstructing these boundaries is crucial for preserving the integrity and accuracy of the building's geometric structure during surface reconstruction. Moreover, in the context of building rendering, well-defined boundaries can significantly enhance the visual quality of the model.

Inspired by $Loss_{rgb}$, boundary loss is estimated by assigning greater weights to the boundaries between various multi-level masks, the detailed formula is as follows:

$$Loss_{boundary} = \alpha \cdot \sum_{(u,v)} w(u,v) \cdot Loss_{L1}(u,v) + \beta \cdot \sum_{(u,v)} w(u,v) \cdot Loss_{ssim}(u,v) \quad (17)$$

where $\alpha$ and $\beta$ are weighting factors. The weight matrix $w(u,v)$ can be referred to Equation 17, in which pixels located at boundaries are with large weight of 10, pixels of masked region are weighed by 1 and non-masked region is not considered during training.

**(2) Multi-level perceptual mask loss**

In addition to emphasizing boundaries, each mask typically contains consistent internal features, such as pixels belonging to the same window or wall surface. Therefore, we introduce a multi-level perceptual masked loss function to optimize the gaussian fields within the masks.

$$Loss_{rgb_k} = \alpha \cdot Loss_{L1_k} + \beta \cdot Loss_{MS-SSIM_k} \quad (18)$$

where $Loss_{L1_k}$ is the L1 loss of $k$-th multi-level mask on an image, $Loss_{MS-SS\ k}$ is the multiscale structural similarity index loss of $k$-th mask on an image.

For $k$-th mask on an image, the masked RGB loss $Loss_{rgb_k}$ is estimated as shown in Equation (18), as well as the mean gradient $grad_{mask_k}$ computed from the corresponding original masked image, its inverse $\delta_i = 1/(grad_{mask_k})$ serves as a weight for the mask during training. Then, by summing the weighted $Loss_{rgb_k}$ of each mask from the multi-level masks, the final multi-level perceptual mask loss can be formulized as:

$$Loss_{mlpm} = \sum_k \delta_k \cdot Loss_{rgb_k} \quad (19)$$

**(3) Total loss**

In addition to the presented two new losses, this paper also employs the depth distortion loss function $Loss_{dept}$ and the normal consistency loss function $Loss_{normal}$ from 2DGS (Huang et al., 2024). Therefore, the final total loss function in this paper is:

$$Loss_{total} = \lambda_1 \cdot Loss_{mlpm} + \lambda_2 \cdot Loss_{boundary} + \lambda_3 \cdot Loss_{depth} + \lambda_4 \cdot Loss_{normal} \quad (20)$$

Where $\lambda_1=0.5, \lambda_2=0.5, \lambda_3=100, \lambda_4=0.05$.

### 3.3 Surface mesh extraction

In this section, based on the trained Masked Gaussian Fields, we first perform multi-directional Screening on the generated Gaussian spheres using the masked points as a reference to accelerate the subsequent processing (Section 3.3.1). Then, in Section 3.3.2, we implement the generation of SDFs and use marching tetrahedra to extract the mesh.

#### 3.3.1 Multi-directional screening of Gaussian spheres

Novel-view images can be rendered just after the training of masked gaussian fields. Next, mesh extraction based on the masked Gaussian fields is required to obtain surface meshes. Conventional mesh extraction methods, such as Marching Cubes (Newman and Yi, 2006) and Poisson reconstruction (Kazhdan and Hoppe, 2013), are ineffective and computationally cost.

GOF (Yu et al., 2024b) improved Tetrahedral mesh extraction and achieved state-of-the-art reconstruction results. Initially, it generates a 3D bounding box at 3-sigma for each Gaussian sphere, followed by constructing tetrahedra based on the Gaussian centers and the corners of the bounding box. Then, this method calculates the opacity and employs a monotonically increasing binary search algorithm to identify level sets, ultimately simplifying the extraction of the mesh. However, GOF exhibits a complexity of $O(N\log N)$, and the increase in the number of tetrahedral vertices and redundancy leads to a rapid increase in computation time. The tetrahedral vertices are derived from the trained Gaussian spheres. Although applying our masks during the training prevents excessive redundant Gaussian spheres from being generated by other objects, gaussian spheres may still occur in the air and at the edges of building due to the ambiguity between rendering and geometry.

To improve the accuracy and speed of surface reconstruction, this paper proposes a precise Gaussian spheres screening strategy based on the masked images. The center points of Gaussian spheres are projected onto the root images. For each center point, we retain those points whose projections lie within the masked regions on all root images. A detailed explanation is provided as follows:

First, according to Equation (21), we project the Gaussian center points $G = \{g_i\}$ onto each masked image $M_j \in M_{root}$, obtaining the projection coordinates $(u_{ij}, v_{ij})$.

$$\begin{pmatrix} u_{ij} \\ v_{ij} \\ 1 \end{pmatrix} = K_j \left( R_j \begin{pmatrix} X_i \\ Y_i \\ Z_i \end{pmatrix} + t_j \right) \quad (21)$$

Then, these Gaussian spheres whose projections of center points fall within the masked regions of the root images are kept as $G'$ in Equation (22), which are indicated by value of 1 on the corresponding root masks.

$$G' = \{ g_i \in G \mid \forall j, M_j(u_{ij}, v_{ij}) = 1 \} \quad (22)$$

The introduced method is expected to reduce the computation load of detecting tetrahedral vertices, speed up the training process, and reduce the occurrence of erroneous polygons in surface reconstruction.

#### 3.3.1 Surface mesh extraction from Masked Gaussian Fields

In this section, we first obtain the opacity of each Gaussian point from the Masked Gaussian Fields, then derive the corresponding SDF values and level sets. Subsequently, we use marching tetrahedra to extract the surface mesh from the level sets.

According to Equation 23, the opacity at any point along the ray is:

$$O_k(G_k, \mathbf{o}, \mathbf{r}, t) = \begin{cases} G_k^{1D}(t) & \text{if } t \leq t^* \\ G_k^{1D}(t^*) & \text{if } t > t^* \end{cases} \quad (23)$$

The accumulated opacity for K Gaussians at any point along the ray can be defined as:

$$O(\mathbf{o}, \mathbf{r}, t) = \sum_{k=1}^{K} \alpha_k \, O_k(G_k, \mathbf{o}, \mathbf{r}, t) \prod_{j=1}^{k-1} \left(1 - \alpha_j O_k(G_k, \mathbf{o}, \mathbf{r}, t)\right) \quad (24)$$

For a 3D point **x**, we define the opacity $O(\mathbf{x})$ as the minimum opacity value among all views and all directions:

$$O(\mathbf{x}) = \min_{(\mathbf{r},t)} O(\mathbf{o}, \mathbf{r}, t) \quad (25)$$

The SDF is computed from the opacity values:

$$SDF(\mathbf{x}) = O(\mathbf{x}) - 0.5 \quad (26)$$

Finally, we use the marching tetrahedra method to extract the surface mesh from the SDF.

## 4. Experiment and Analysis

To validate the efficacy of the proposed method, extensive experiments are conducted on two public benchmarks utilizing several common evaluation metrics (see section 4.1). In general, the primary goal of our experiments is to demonstrate the superior performance of our MGFs in surface reconstruction. Additionally, the results of novel view rendering, as a byproduct of our method, are investigated as well. Section 4.2 compares the rendering results with several state-of-the-art gaussian-based rendering methods. The next section presents surface reconstruction results from various volume rendering-based methods and a traditional method implemented by COLMAP. In section 4.4, we perform comprehensive ablation studies to elucidate how each component of our method contributes to achieving optimal results.

### 4.1. Dataset and evaluation metrics

#### 4.1.1 Datasets and settings

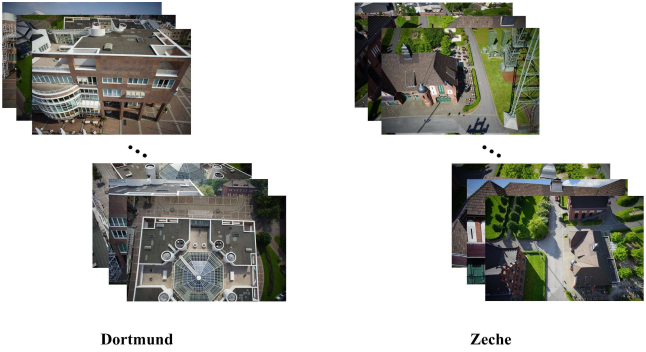

Fig. 2 Sample images of *Dortmund* and *Zeche*

This paper utilizes two public UAV image datasets, *Dortmund* and *Zeche*, which are from the ISPRS Multi-platform Benchmark (Nex et al., 2015) The *Dortmund* dataset was collected via a Sony Nex-7 mounted on a multi-rotor UAV (model DJI S800), flying over the area around the *Dortmund* City Hall. This dataset includes 146 images with a resolution of 6000×4000 pixels. The *Zeche* dataset contains 147 images collected by the same sensors at *Zeche Zollern* in *Dortmund*. Both datasets focus on building scenarios with some background noise, and two sample images are shown in Fig. 2.

All the reported experiments were conducted on a machine equipped with Intel(R) Xeon(R) Gold 6133 CPU @ 2.50GHz with 20 cores and a single RTX 4090 GPU with 24GB of memory. The ground truth (GT) mesh for the two datasets used in this paper was reconstructed by Zhu et al. (2020). All tested images were downsampled to 750x500 pixels to fulfil the requirement of our hardware.

#### 4.1.2 Evaluation metrics

The masked Gaussian fields are capable of both surface reconstruction and novel-view rendering, and both aspects were comprehensively evaluated in our experiments.

First, to assess novel view image rendering, three widely-used evaluation metrics are applied: Peak Signal to Noise Ratio (*PSNR*) (Chen and Wang, 2024), Structural Similarity Index (*SSIM*) (Lin et al., 2024), and Learned Perceptual Image Patch Similarity (*LPIPS*) (Zhang et al., 2018). However, for a fair comparison, only the masked region is investigated. Specifically, both the ground truth and rendered images were masked according to section 3.1.2, retaining only the pixel values in the masked regions and other regions set to zero. The evaluation metrics are as follows:

$$PSNR = 10 \times \log_{10}\left(\frac{MAX^2}{MSE}\right) \quad (27)$$

MAX represents the maximum pixel value of the image. MSE is the

$$SSIM(x,y) = \frac{(2\mu_x\mu_y + c_1)(2\sigma_{xy} + c_2)}{(\mu_x^2 + \mu_y^2 + c_1)(\sigma_x^2 + \sigma_y^2 + c_2)} \quad (28)$$

$x$ and $y$ represent the original and rendered images. $\mu_x$ and $\mu_y$ are the mean luminance of images $x$ and $y$. $\sigma_x^2$ and $\sigma_y^2$ are the corresponding variances (contrast). $\sigma_{xy}$ is the covariance between images $x$ and $y$. The larger *SSIM* is, the structural similarity better.

$$LPIPS(x,y) = \sum_l \frac{1}{H_l W_l} \sum_{h,w} \| \phi_l(x)_{h,w} - \phi_l(y)_{h,w} \|_2^2 \quad (29)$$

$\phi_l$ represents the feature extraction function of the $l$ layer (e.g., feature map of a pre-trained convolutional neural network). $H_l$ and $W_l$ are the height and width of the $l$ layer feature map. $\phi_l(x)_{h,w}$ and $\phi_l(y)_{h,w}$ are the feature vectors at position $(h,w)$ of the $l$ layer feature map for images $x$ and $y$, respectively.

Secondly, for the evaluation of surface reconstruction, in this work, we refer to Mazzacca et al. (2023) and Yan et al. (2023). Specifically, we first sample two point clouds from the reconstructed mesh and the GT mesh, then calculate the corresponding Accuracy, Completeness, and *F1* score. Accuracy measures the precision of the generated surface, taking GT mesh as reference. Completeness indicates the number of points in the GT are reconstructed using the proposed method. *F1* combines Accuracy and Completeness, providing a comprehensive evaluation of the surface reconstruction results. These three metrics are computed as follows:

$$ACCURACY = \frac{\sum_{i=1}^{S}(DisT_i < Th)}{S} \quad (30)$$

$$COMPLETENESS = \frac{\sum_{i=1}^{T}(DisS_i < Th)}{T} \quad (31)$$

$$F1 = 2 \cdot \frac{ACCURACY \cdot COMPLETENESS}{ACCURACY + COMPLETENES} \quad (32)$$

where $DisT$ represents the point-to-point distance from the obtained mesh to the corresponding points in the GT mesh. $DisS$ represents the distance in the opposite direction, $S$ and $T$ are the total number of points in the obtained mesh and the GT mesh, respectively, and $Th$ is the threshold set to filter out out-of-range points, which set as 0.05m in this paper.

## 4.2. Novel view rendering results

In this section, we compare our proposed method with four state-of-the-art Gaussian-based novel view rendering methods, including 3DGS (Kerbl et al., 2023), SuGaR (Guédon and Lepetit, 2024), 2DGS (Huang et al., 2024) and GOF (Yu et al., 2024b), using the *Dortmund* and *Zeche* dataset. All four methods follow their default settings, using the original entire images for training, and the number of training iterations is set to 30,000 for all methods, including our MGFs. For a fair comparison with our MGFs, only the corresponding masked building regions are displayed after rendering and applied for computing *PSNR*, *SSIM*, and *LPIPS*.

### 4.2.1 Dortmund

As depicted in Table 1, for all evaluation metrics, we achieve the best. Notably, we surpassed *PSNR* of 32, while the GOF follows, slightly exceeding 30. In contrast, all other methods scored below 30. This indicates that our method excels structural similarity and other relevant aspects, thereby demonstrating the superiority of the proposed MGFs. 3DGS achieved the second-best *SSIM* (0.957) and *LPIPS* (0.022) values, and the third-best *PSNR* (29.958) value. This can be attributed to fact that it does not optimize the distribution of 3D Gaussian spheres for mesh extraction by design, which facilitates precise color synthesis for generating novel images. GOF surpasses 3DGS in *PSNR* while closely rivaling it in *SSIM* (0.952) and *LPIPS* (0.025) due to its enhanced densification strategy and interaction of Gaussian opacity fields. Conversely, 2DGS, instead of 3D Gaussian spheres, employs directional 2D Gaussian discs to focus more on 2D planes, resulting in less flexibility on color rendering, and it is just superior to SuGaR that concentrates on surface reconstruction by aligning the Gaussian spheres to the real scene and reduces synthesis of pixel color.

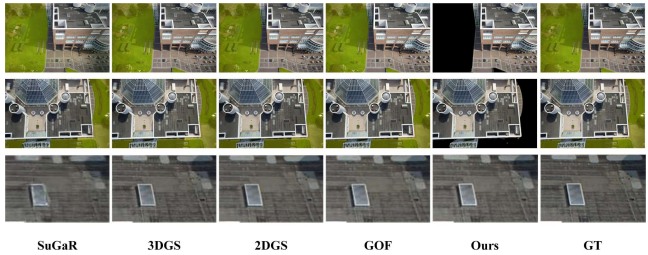

Fig. 3 Rendering results of *Dortmund*. The first and second row are two different rendering results obtained by various methods, and the third row is a zoom-in display of the rendering results. The light-green background indicates the non-building area.

Table 1 *PSNR, SSIM, LPIPS* results of several methods on *Dortmund*. Bests are highlighted in bold.

| Method | PSNR ↑ | SSIM ↑ | LPIPS ↓ |
|---|---|---|---|
| 3DGS | 29.958 | 0.957 | 0.022 |
| SuGaR | 23.541 | 0.843 | 0.082 |
| 2DGS | 29.799 | 0.944 | 0.033 |
| GOF | 30.383 | 0.952 | 0.025 |
| **Ours** | **32.121** | **0.966** | **0.016** |

Fig. 3 shows the rendering results of two example scenes from the *Dortmund*. As shown in the fifth column of Fig. 3, our MGF only renders the building regions, while other methods render the entire scene. Since our MGF employs masks for light filtering, training is focused solely on the masked building areas. Looking at the zoom-in images in the third row, we can see that the proposed MGFs closely resembles ground truth (GT) images in terms of fidelity. In contrast, images rendered by 3DGS resemble those from 2DGS and GOF, albeit with slightly reduced clarity compared to MGFs. Lastly, images produced by SuGaR exhibit the least sharpness and clarity among all compared methods, consistent with the results listed in Table 1. In summary, our method, which exclusively targets the building area and proposes the corresponding loss functions and rendering method, effectively achieves a commendable balance among structural similarity, local consistency, and fidelity.

### 4.2.2 Zeche

As shown in Table 2, in general, the quantitative rendering results for *Zeche* basically exhibit a trend similar to those for *Dortmund* presented in Table 1. However, it is particularly notable that, compared to other methods, the additional improvement in PSNR values obtained by our method on *Zeche* is larger than the improvement on *Dortmund*. For example, on *Zeche*, the *PSNR* values of MGFs (36.903) is 5.428 higher than the 3DGS (31.475), whereas on *Dortmund*, *PSNR* values of MGFs (32.121) is only 2.163 higher than the 3DGS (29.958). As illustrated in Fig. 3, MGFs also achieves rendering results that are closest to the ground truth (GT).

These performances can be attributed not only to inherent properties of various methods explained in Section 4.2.1, but also to the distinct characteristics of these two datasets. The *Dortmund* dataset was collected by a drone flying at a lower altitude around building, while the *Zeche* dataset was at a higher altitude. The higher altitude during image capturing in the *Zeche* dataset results in the target building occupying a smaller proportion of the images, which could increase the influence of the non-building regions during training and rendering.

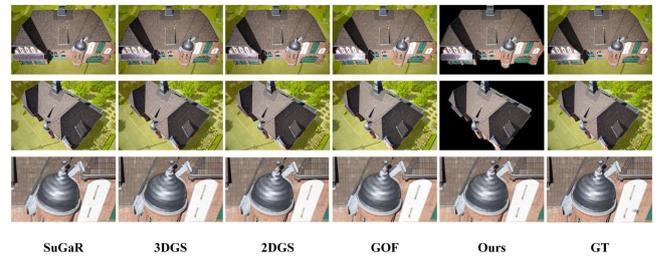

Fig. 4 Rendering results of *Zeche*. The first and second rows of images are two different rendering results obtained by various methods, and the third row is a zoom-in display of the rendering results. The light-green background indicates the non-building area.

Table 2 *PSNR, SSIM, LPIPS* results of several methods on *Zeche*. Bests are highlighted in bold.

| Method | PSNR ↑ | SSIM ↑ | LPIPS ↓ |
|---|---|---|---|
| 3DGS | 31.475 | 0.953 | 0.022 |
| SuGaR | 28.301 | 0.896 | 0.047 |
| 2DGS | 31.050 | 0.944 | 0.031 |
| GOF | 30.943 | 0.945 | 0.029 |
| **Ours** | **36.903** | **0.978** | **0.013** |

## 4.3. Surface reconstruction results

To demonstrate the advancements in reconstructing surface meshes, we compare three SOTA methods based on Gaussian fields: 2DGS (Huang et al., 2024), SuGaR (Guédon and Lepetit, 2024), and GOF (Yu et al., 2024b), along with one NeRF-based method, Nerfacto (Tancik et al., 2023), and one traditional method based on COLMAP (Schonberger and Frahm, 2016). Adhering to the default settings of these methods, we set the number of training iterations to 30,000, and the results after 7000 training iterations in GOF are also reported. However, for our MGFs, 7,000 training iterations are sufficient to achieve optimal reconstruction results (see more details in our ablation studies in section 4.4), which are reported in this section.

### 4.3.1 Dortmund

Fig. 5 shows the surface reconstruction results of *Dortmund* using various methods. The first column is the original output meshes, the second column shows the reconstructed surface meshes of the target building, and the third and fourth columns provide zoom-in views of two local places from the building. Overall, our MGFs achieves the best reconstruction results, focusing more on the building itself with clean and explicit contours. In contrast, other methods tend to reconstruct noisier areas.

Based on the reconstructed details depicted in the last two columns of Fig.5, we can find that both COLMAP and Nerfacto reconstruct the complex hollow steel frame by a highly simplified mesh block. SuGaR generates numerous irregular protrusions and depressions, which are particularly noticeable on walls and roofs. This can be stemmed from the estimated inaccurate depth map using a loose regularization. 2DGS exhibits a pronounced over-smoothing reconstruction, especially evident in the areas with red dotted boxes, such as the omission of the complex steel frame and the presence of holes in some flat regions. This can be explained by the fact that the 2D directional Gaussian discs are more prone to mesh 2D planes and inferior at constructing 3D complex details. Comparing GOF_7000 (7000iterations) and GOF_30000 (30000 iterations), the former produces significant noise due to the lack of geometric regularization, and it is improved by GOF_30000 integrating the constraints of depth distortion and normal consistency (as done in 2DGS), but there are still some noisy areas and holes due to the influence of non-building area. In comparison to the ground truth, our GMFs can generate the most closed surface reconstruction. Specifically, planar regions, such as windows and walls without any holes, are successfully meshed. Additionally, MGFs are capable of reconstructing complex structures, such as steel frameworks with least noise. This improvement could be attributed to the proposed two new loss functions and masked Gaussian fields that concentrate on building areas.

For quantitative evaluation, we sample 1 million 3D points on target building from GT mesh and generated mesh, respectively. Note that the GT mesh itself is not accurately reconstructed on steel frameworks, thus they are not considered in this quantitative evaluation. Three evaluation metrics and cost time of various methods are provided in Table 3, it can be found that our method achieves either the best or second-best results. It is important to note that, except Nerfacto that applies NeRF-based surface reconstruction, all other methods achieve a relatively high accuracy (above 98.0). Nevertheless, using point cloud distances for numerical analysis cannot fully reflect the reconstruction quality, for instance, SuGaR, showing noisy reconstruction (as Fig. 5 shows), samples points distributed around the GT points, resulting in a high accuracy score. Hence, looking at completeness and *F1* score, our method is only inferior to 2DGS on accuracy by just 0.1%. However, 2DGS fails to reconstruct the steel frameworks.

For time efficiency, our method also demonstrates significant advantages, outperforming most methods and being considerable to the fastest method, 2DGS. In addition, we would like to remind that only an extra 1 min is needed for generating masked information.

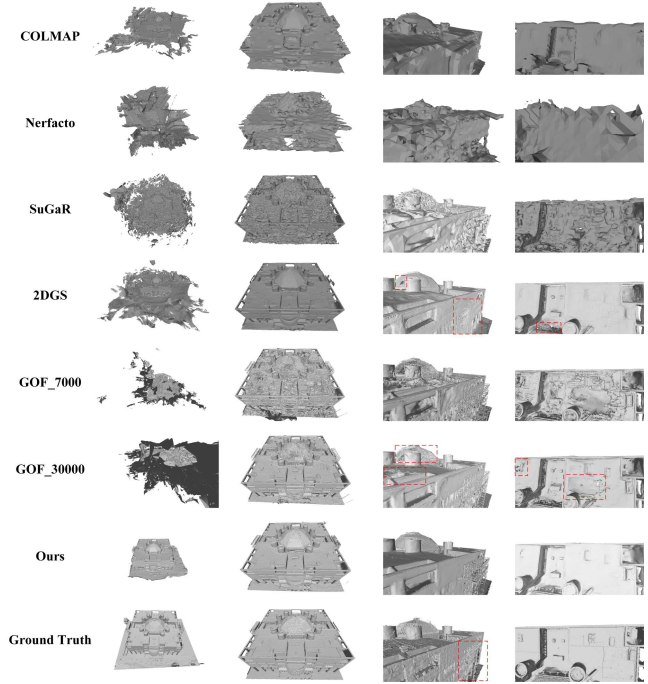

Fig. 5 Surface reconstruction results of *Dortmund*. The first column images are the overall meshes of various methods, the second column images denote the building surface meshes. The third and fourth column images are zoom-in displays of two local places from the building

Table 3 *Accuracy*, *Completeness*, *F1*(in %), cost time of various methods on *Dortmund*. Bests and second bests are highlight in bold and red, respectively.

| Method | Accuracy | Completeness | F1 | time |
|---|---|---|---|---|
| COLMAP | 87.6 | 98.6 | 92.8 | 0.5h |
| Nerfacto | 54.3 | 66.4 | 59.7 | 1 h |
| SuGaR | 85.4 | 99.8 | 92.1 | 3 h |
| 2DGS | **98.7** | 99.2 | 98.9 | **0.3 h** |
| GOF_7000 | 70.9 | 99.6 | 83.0 | 0.8 h |
| GOF_30000 | 73.8 | **99.9** | 84.9 | 2.3 h |
| **Ours** | 98.6 | **99.9** | **99.2** | 0.4h+(1min) |

### 4.3.2 Zeche

Fig. 6 illustrates the surface reconstruction results of various methods on *Zeche*. Akin to the results shown in Fig. 5, the surfaces reconstructed by COLMAP and Nerfacto are relatively rough. SuGaR exhibits significant noise, while 2DGS's results lack some detailed components, with parts of the walls and roof missing. GOF exhibits severe noise after 7,000 iterations; although it reconstructs all building structures after 30,000 iterations, considerable noise remains, such as depressions in smooth areas like facades and roof decorations. The proposed MGFs, on the other hand, produces meshes that most closely resemble the GT mesh.

Table 4 *Accuracy*, *Completeness*, *F1*(in %), cost time of various methods on *Zeche*. The Bests and second-bests are highlighted in bold and red, respectively.

| Method | *Accuracy* | *Completeness* | *F1* | time |
|---|---|---|---|---|
| COLMAP | 69.5 | 70.3 | 69.9 | 0.5h |
| Nerfacto | 32.4 | 26.7 | 29.3 | 1 h |
| SuGaR | 43.4 | 78.0 | 55.8 | 3 h |
| 2DGS | **89.3** | 60.5 | 72.1 | **0.3 h** |
| GOF_7000 | 67.8 | 78.8 | 72.8 | 0.8 h |
| GOF_30000 | 77.2 | 85.1 | 81.0 | 2.3 h |
| **Ours** | 87.4 | **90.5** | **88.9** | **0.3h+(1min)** |

Table 4 presents the quantitative evaluation results. Similar to the setup in *Dortmund*, we sampled 1,000,000 points on both the GT mesh and the generated mesh. MGFs achieved superior results in completeness, accuracy, and *F1* score, demonstrating its effectiveness. Notably, the Accuracy indicator of the 2DGS method is slightly higher than ours. This is because many polygons of the wall and roof missed in the 2DGS mesh are not considered when computing Accuracy between 2DGS and GT mesh, leading to the higher accuracy of 2DGS. However, these polygons are mostly reconstructed by MGFs, albeit not very accurately, as MGFs only use UAV images, whereas the GT mesh is reconstructed using both UAV and terrestrial images.

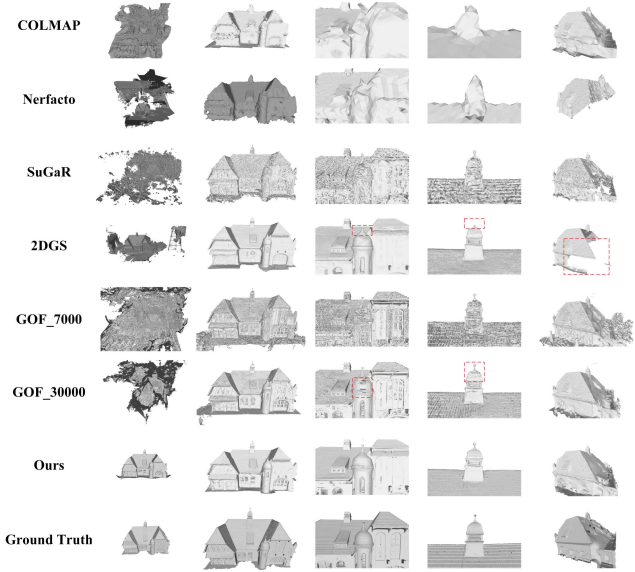

Fig. 6 Surface reconstruction results of Zeche. The first column images are the overall meshes of various methods, the second column images denote the building surface meshes. The last three column images are zoom-in displays of three local places from the building.

### 4.4. Ablation study

To evaluate the effectiveness of each step in the proposed method (as illustrated in Fig. 1), we conduct ablation studies on both rendering and surface reconstruction using Dortmund. The corresponding experimental settings are as follows:

(a) *Ours w/o masked image*. The entire image is treated as the region of interest for training.

(b) *Ours w/o masked points*. Instead of the masked points, the original initial points are used as input.

(c) *Ours w/o multi-level perceptual mask loss and boundary loss*. The multi-level perceptual mask loss and boundary loss proposed in this paper are excluded.

(d) *Ours w/o multi-directional screening of gaussian spheres*. multi-directional screening of Gaussian spheres strategy is not used before mesh extraction.

(e) *Ours_7000*. All above improvements with masked information are applied with 7000 training iterations for surface extraction.

(f) *Ours_30000*. Similar to (e), but the training iteration is set to 30000 for surface extraction.

Table 5 presents the numerical rendering results of the ablation experiments (a), (b), (c) and our MGFs (Ours). The most significant impact on rendering is observed when the setting of *w/o multi-level perceptual mask loss and boundary loss* is switched off, resulting in a *PSNR* reduction of 3.489 comparing to MGFs. These two novel loss functions enhance the information of edge and masked building area, thereby increasing local consistency and significantly improving the rendering quality. Additionally, *w/o masked points* led to a *PSNR* reduction of 1.185, because masked 3D points can reduce interference from redundant point clouds in non-building areas, and improves overall rendering efficiency. Furthermore, *Ours w/o mask image* resulted in a *PSNR* decrease of 1.113 due to that the detected masks to local building optimization, thereby improving rendering accuracy.

Fig. 7 illustrates the surface reconstruction results of various ablation experiments. In particular, *Ours w/o mask image* leads to an excessively large reconstruction, including a significant amount of noisy background areas. Additionally, the absence of masks yields small holes. *Ours w/o masked points* results in the appearance of extraneous polygons around the mesh and reduces the smoothness of the surface of mesh. *Ours w/o multi-level perceptual mask loss and boundary loss* leads to irregular mesh on the surface and confusion between polygons from neighboring regions. *Ours w/o Multi-Directional Screening of Gaussian spheres* generates some unexpected redundant polygons outside the building. In the last, *Ours w 30000 training iterations* yields a mesh that is relatively similar to the results obtained with MGFs.

Table 5 *PSNR*, *SSIM*, *LPIPS* results of ablation studies on *Dortmund*.

| Method | *PSNR* | *SSIM* | *LPIPS* |
|---|---|---|---|
| (a) | 31.008 | 0.961 | 0.021 |
| (b) | 30.936 | 0.962 | 0.020 |
| (c) | 28.632 | 0.955 | 0.024 |
| Ours | 32.121 | 0.966 | 0.016 |

Table 6 provides a quantitative assessment of ablation studies on reconstruction results. Comparing to ours_7000, all the other ablation experiments exhibit a deterioration across all evaluation metrics. Similar to Table 5, the most significant decrease is observed when the two new losses are not utilized. When comparing the reconstruction results after 7,000 and 30,000 iterations, it can be seen that the latter requires more than four times the computational time of the former, yet achieves very similar metrics. Consequently, it is deemed suitable to extract the mesh after 7,000 iterations for our MGFs.

Table 6 *Accuracy*, *Completeness*, *F1*(in %) and cost time of ablation methods on the *Dortmund*.

| Method | Accuracy | Completeness | F1 | time |
|---|---|---|---|---|
| (a) | 96.8 | 99.9 | 98.3 | 0.5 h |
| (b) | 97.7 | 99.9 | 98.8 | 0.4 h |
| (c) | 93.3 | 99.8 | 96.5 | 0.4 h |
| (d) | 98.5 | 99.3 | 98.9 | 0.4 h |
| Ours_7000 | 98.6 | 99.9 | 99.2 | 0.4 h |
| Ours_30000 | 98.7 | 99.9 | 99.2 | 1.7 h |

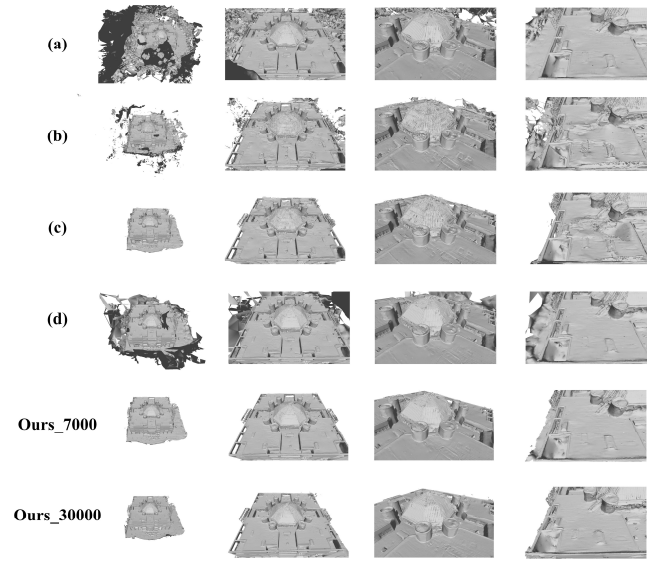

Fig. 7 Results of ablation experiments on surface reconstruction on *Dortmund*. The first column of images are the overall mesh of the various settings, the second column of images shows surface meshes of target building. The third and fourth columns are zoom-in views of two local places from the building.

| | view1 | view2 | Accuracy | Completeness | F1 | time |
|---|---|---|---|---|---|---|
| COLMAP (750*500) | | | 87.6 | 98.6 | 92.8 | 0.5h |
| COLMAP (1500*1000) | | | 87.1 | 99.7 | 93.0 | 1.2 h |
| Ours (750*500) | | | 98.6 | 99.9 | 99.2 | 0.4 h |
| Ours (1500*1000) | | | 98.8 | 99.9 | 99.3 | 0.5h |

Fig. 8 Surface reconstruction results of images with different resolutions applied in COLMAP and our MGFs. The first column shows the method used and the resolution of the image, the second and third columns show the detailed display of the building meshes, and the remaining four columns are the Accuracy, Completeness, F1 and time of the meshes.

Furthermore, to investigate the influence of image resolution on surface reconstruction, we doubled the resolution for the input images. As Fig. 8 shows, for low-resolution images, COLMAP produces poor geometric reconstruction results, capturing only rough outlines of the building. These results are qualitatively improved when using high-resolution images, as more details are revealed; however, the processing time increases by 2.4 times. Despite this improvement, COLMAP with high-resolution images still lags behind our method using low-resolution images in terms of both surface reconstruction accuracy and time efficiency. When high-resolution images are input into our method, the processing time increases by 1.2 times, but the improvement in surface reconstruction is limited (only the *accuracy* is improved by 0.2%).

## 5. Conclusion

In this work, we present a novel masked Gaussian Fields (MGFs) for surface reconstructing of the building using multi-view images. First, EfficientSAM and COLMAP are employed to obtain multi-level masks and masked points of building regions. Next, we presented a Masked Gaussian Fields based on boundary ray enhancement and masked rays inside the building. In addition, a novel boundary loss is proposed by the predicted values of boundary ray using a new weighted volume rendering, as well as a multi-level perceptual masked loss, encompassing all the pixels belonging to the detected multi-level masks. Finally, we improve the tetrahedron extraction method based on the masked gaussian spheres and multi-directional filtering for building surface mesh extraction.

Extensive experimental results demonstrate that, compared to the traditional pipeline of COLMAP and several SOTA gaussian-based surface reconstruction and novel-view rendering methods, our MGFs can yield more accurate and detailed meshes for building in a time-efficient manner. In the future, we would like to extend our method to deal with large scenarios containing multiple different building.

## CRediT authorship contribution statement

**Tengfei Wang**: Writing – review & editing, Writing – original draft, Visualization, Validation, Software, Methodology, Investigation, Formal analysis, Data curation. **Xin Wang**: Writing – review & editing, Methodology, Conceptualization, Project administration, Supervision. **Linxia Ji**: Writing – review & editing. **Rui Xia:** Data curation. **Zongqian Zhan**: Writing – review & editing, Funding acquisition , Supervision.

## Acknowledgements

This work was supported by the National Natural Science Foundation of China (No.42301507) and Natural Science Foundation of Hubei Province, China (No. 2022CFB727).